\documentclass[letterpaper, 10 pt, conference]{ieeeconf}

\IEEEoverridecommandlockouts

\usepackage{dblfloatfix}

\usepackage{color, colortbl}
\usepackage[colorlinks,bookmarksnumbered,citecolor=orange,urlcolor=orange]{hyperref}
\usepackage{graphicx}
\graphicspath{{Figures}}
\DeclareGraphicsExtensions{.pdf,.png}
\usepackage{bigstrut}
\usepackage[english]{babel}

\usepackage{csquotes}

\usepackage[T1]{fontenc}
\usepackage{url}
\usepackage{multirow}
\usepackage{float}
\usepackage{xcolor}
\usepackage{hyperref}
 \hypersetup{
     colorlinks=true,
     linkcolor=orange,
     filecolor=orange,
     citecolor=orange,      
     urlcolor=orange,
     }

\let\labelindent\relax
\usepackage{graphicx}
\usepackage{amsmath}
\usepackage{enumitem}

\usepackage{float}   
\usepackage[ruled,vlined]{algorithm2e}
\usepackage{epstopdf} 
\usepackage{arydshln}

\usepackage{epsfig} % for postscript graphics files
\usepackage{times} % assumes new font selection scheme installed
\usepackage{amssymb}  % assumes amsmath package installed

\usepackage{booktabs}
\usepackage{pifont}
\newcommand{\cmark}{\ding{51}} % checkmark
\newcommand{\greencmark}{\textcolor{green}{\ding{51}}}
\newcommand{\xmark}{\ding{55}} % crossmark

\usepackage{cite}

\usepackage{glossaries}
\usepackage{multicol}
\usepackage{comment}

\usepackage[font={small}]{caption} % Optional: makes captions smaller
\usepackage{fancyhdr}
\title{\LARGE
IndoorR2X: Indoor Robot-to-Everything Coordination\\ with LLM-Driven Planning
\vspace{-0.5em}
}

\author{
    Fan Yang$^{1}$, Soumya Teotia$^{2}$, Shaunak A. Mehta$^{1}$, Prajit KrisshnaKumar$^{1}$, Quanting Xie$^{2}$, \\ 
    Jun Liu$^{2}$, Yueqi Song$^{2}$, Wenkai Li$^{2}$, Atsunori Moteki$^{1}$, Kanji Uchino$^{1}$, and Yonatan Bisk$^{2}$
    \thanks{$^{1}$Fujitsu Research}
    \thanks{$^{2}$Carnegie Mellon University}
}
\begin{document}
\maketitle
\thispagestyle{fancy}

%%%%%%%%%%%%%%%%%%%%%%%%%%%%%%%%%%%%%%%%%%%%%%%%%%%%%%%%%%%%%%%%%%%%%%%%%%%%%%%%
\begin{abstract}

Although robot-to-robot (R2R) communication improves indoor scene understanding beyond what a single robot can achieve, R2R alone cannot overcome partial observability without substantial exploration overhead or scaling team size. In contrast, many indoor environments already include low-cost Internet of Things (IoT) sensors (e.g., cameras) that provide persistent, building-wide context beyond onboard perception. We therefore introduce IndoorR2X, a benchmark and simulation framework for Large Language Model (LLM)-driven multi-robot task planning with Robot-to-Everything (R2X) perception and communication in indoor environments. IndoorR2X integrates observations from mobile robots and static IoT devices to construct a global semantic state that supports scalable scene understanding, reduces redundant exploration, and enables high-level coordination through LLM-based planning. IndoorR2X provides configurable simulation environments, sensor layouts, robot teams, and task suites to systematically evaluate semantic-level coordination strategies. Extensive experiments across diverse settings demonstrate that IoT-augmented world modeling improves multi-robot efficiency and reliability, and we highlight key insights and failure modes for advancing LLM-based collaboration between robot teams and indoor IoT sensors. Project page: \url{https://fandulu.github.io/IndoorR2X_project_page/}.

\end{abstract}
%\vspace{-0.5em}
%%%%%%%%%%%%%%%%%%%%%%%%%%%%%%%%%%%%%%%%%%%%%%%%%%%%%%%%%%%%%%%%%%%%%%%%%%%%%%%%
\section{INTRODUCTION}

Indoor service robots are transitioning from single-agent demos to \emph{teams} that must jointly carry out long-horizon tasks such as cleaning, cooking assistance, object fetching, and device operation~\cite{kant2022housekeep, kannan2024smart, zhang2025lamma, gosrich2025online}. In realistic homes and offices, however, multi-robot coordination is fundamentally constrained by \emph{partial observability}: each robot only sees what lies in its current field of view and what it has already explored. Under these constraints, teams frequently waste effort through redundant exploration, inconsistent beliefs about object locations or device states, and brittle task allocation when plans must be revised online.
\begin{figure}[th!]
  \centering
  \includegraphics[width=\linewidth]{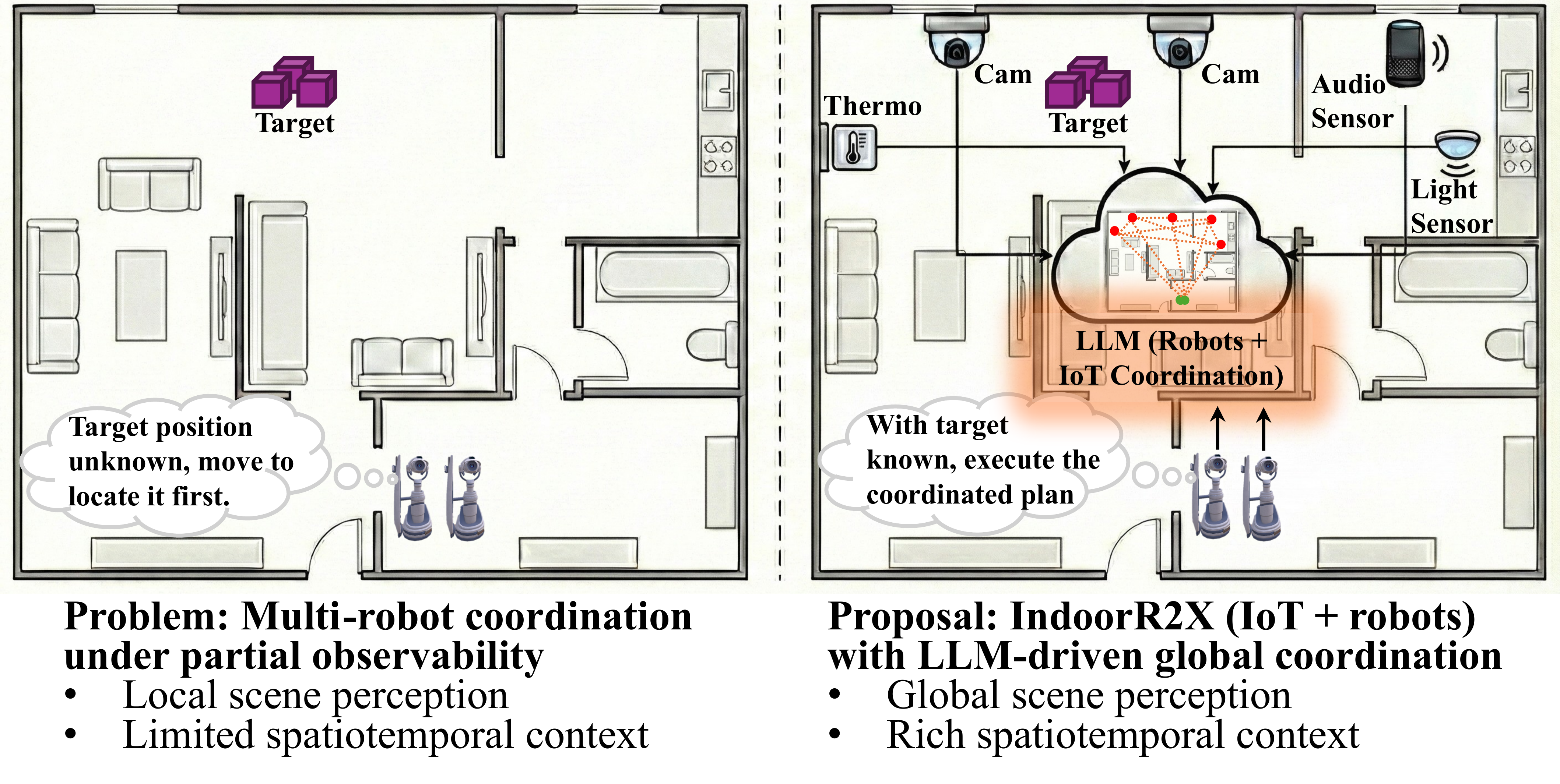}
  \vspace{-.8em}
  \caption{\textbf{Motivation for IndoorR2X.} Augmenting robot perception with global IoT context via LLMs for efficient coordination.}
  \label{fig:demo}
\end{figure}

At the same time, indoor environments are increasingly instrumented with ambient IoT sensors, which can provide persistent, wide-coverage observations unavailable to any single robot~\cite{sabri2018integrated, wilson2019robot,romeo2020internet, aroulanandam2022sensor}. Despite this opportunity, most existing LLM-driven multi-robot frameworks either (i) implicitly assume oracle-level access to global scene state in simulation or (ii) focus primarily on robot-to-robot communication without systematically modeling how heterogeneous IoT sensing can be fused into a shared state representation for planning. This leaves an open question: \emph{How can an indoor robot fleet leverage LLM-based reasoning to coordinate reliably under partial observability while exploiting existing IoT sensing to reduce redundant exploration and planning costs?}
\begin{table*}[t]
\centering
\footnotesize
\caption{\textbf{Comparison of \textit{IndoorR2X} to representative benchmark families.} ``X'' denotes the external sensing augmentation beyond multi-agent on-board sensing. ``LLM Coord.'' indicates whether LLMs are used for multi-agent coordination.}
\label{tab:benchmark_comparison}
\setlength{\tabcolsep}{4pt}
\begin{tabular}{lcccc}
\toprule
\textbf{Benchmark Family} & \textbf{Scene}  & \textbf{Multi-Agent} & \textbf{X (IoT sensors)} & \textbf{LLM Coord.} \\
\midrule
\textbf{(A)} V2X-SIM~\cite{li2022v2x}, V2X-REAL~\cite{xiang2024v2x}& Outdoor  & \cmark & \cmark & \xmark \\
\textbf{(B)} V2X-LLM~\cite{wu2025v2x}, AccidentGPT~\cite{wang2023accidentgpt} & Outdoor  & \cmark & \cmark & \cmark \\
\textbf{(C)} TWOSTEP~\cite{bai2024twostep}, SMART-LLM~\cite{kannan2024smart}, HeterCol~\cite{liu2024heterogeneous}, EMOS~\cite{chen2025emos}  & Indoor  & \cmark & \xmark & \cmark \\
\textbf{IndoorR2X (ours)} & Indoor  & \greencmark & \greencmark & \greencmark\\
\bottomrule
\end{tabular}
\end{table*}
We argue that addressing this question requires two ingredients: (1) a benchmark that explicitly enforces realistic perception limits so that no agent is omniscient, and (2) a framework that can integrate heterogeneous observations into a unified representation that supports online multi-agent planning. To this end, we introduce \textit{IndoorR2X}, the first benchmark and simulation framework for evaluating \emph{LLM-powered multi-robot task planning and execution} in indoor Robot-to-Everything (R2X) settings. 
\textit{IndoorR2X} consists of 85 multi-room environments that provide the scale necessary to support complex household tasks involving joint navigation and manipulation, as well as navigation-only tasks. 
We enforce realistic partial observability by restricting each robot's knowledge to its immediate field of view and visited areas. This constraint renders global coordination non-trivial, directly motivating the use of IoT sensors as a critical supplementary information source.

\textit{IndoorR2X} is paired with a coordination framework centered around a \textbf{coordination hub} that maintains a \textbf{global semantic state} by aggregating observations from both mobile robots and static IoT devices (the ``X'' in R2X). An LLM operates as an \emph{online planner} over this shared state, producing a parallelizable plan represented as a dependency graph, while a system orchestrator executes actions, monitors outcomes, updates the world model, and triggers replanning upon failures. This design enables dynamic coordination under evolving, incomplete information.

Our experiments systematically isolate the roles of information sharing and IoT sensing in multi-robot coordination. Comparing isolated robots (IR), robot-to-robot sharing (R2R), and full robot-to-everything integration (R2X), we find that inter-robot communication is critical for success under partial observability, while IoT-augmented world modeling further reduces action steps, path length, and LLM token cost without sacrificing success. We additionally show that coordination quality depends strongly on LLM capability, incurs increasing overhead as team size grows, and is robust to missing IoT signals but more sensitive to semantic misinformation (e.g., incorrect device states).

Our contributions are threefold:
\begin{itemize}[leftmargin=*, topsep=0pt, itemsep=0.2em, parsep=0pt]
    \item \textbf{Novel R2X Benchmark:} We introduce \textit{IndoorR2X}, the first indoor multi-robot benchmark that strictly enforces partial observability and integrates configurable IoT sensors to evaluate realistic team coordination.
    \item \textbf{LLM-Driven Semantic Fusion:} We propose a centralized framework that fuses onboard robot perception with ambient IoT signals into a shared global semantic state, enabling LLMs to plan parallel tasks without exhaustive physical exploration.
    \item \textbf{Empirical \& Real-World Validation:} Extensive simulations and physical deployments show that our framework reduces path length, action steps, and LLM token costs, while exhibiting high resilience to missing sensor data.
\end{itemize}

\begin{figure*}[th!]
  \centering
  \includegraphics[width=0.95\linewidth]{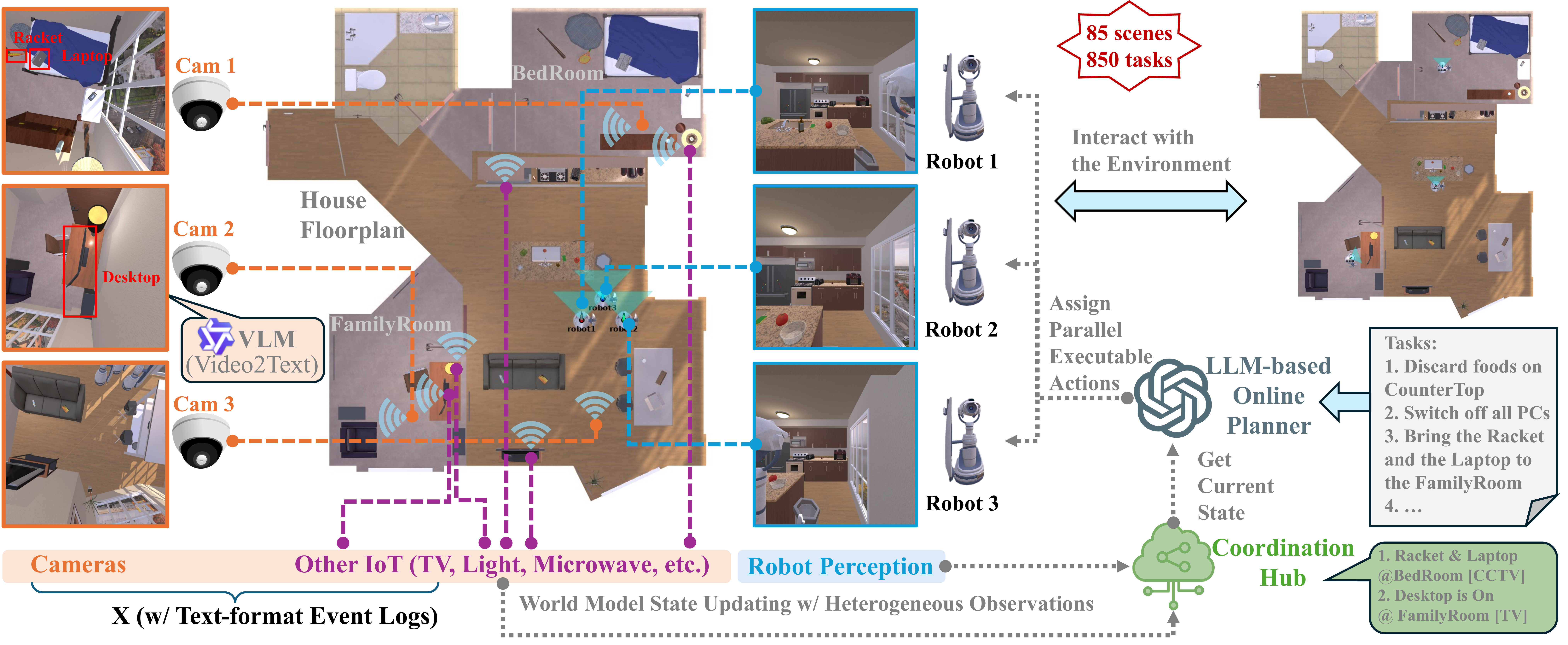}
  \vspace{-0.3em}
  \caption{\textbf{Our IndoorR2X framework.} CCTV observations and other IoT device signals are collected to augment the world model beyond the perception range of the robots’ ego cameras. These heterogeneous observations are synchronized through a coordination hub, where an LLM-based online planner generates parallel actions for each robot and executes them to perform their respective tasks. As an example scenario, robots are assigned to perform household tasks in the morning. After potential overnight changes to object locations or device statuses (e.g., TVs), robots first update their indoor world model by leveraging the ``X'' observations.}
  \label{fig:framework}
\end{figure*}

\section{RELATED WORK}
\label{sec:related_work}

Table~\ref{tab:benchmark_comparison} positions \textit{IndoorR2X} relative to three prior benchmark families. We focus this section on the two technical gaps that motivate our benchmark and framework: (i) how to exploit infrastructure/IoT sensing for indoor embodied coordination, and (ii) how to evaluate LLM-based multi-robot planning under realistic partial observability.

\subsection{IoT-Augmented Perception}
Cooperative perception with infrastructure support has been extensively studied in autonomous driving under the Vehicle-to-Everything (V2X) umbrella~\cite{yusuf2024vehicle, v2x_1}. Benchmarks such as V2X-SIM~\cite{li2022v2x} and V2X-REAL~\cite{xiang2024v2x} formalize how roadside sensing can be shared to improve detection and tracking beyond a single vehicle’s view. Recent work also explores integrating large models into V2X pipelines for higher-level understanding and analysis~\cite{wang2023accidentgpt, mahmud2025integrating, wu2025v2x}.

However, these outdoor settings primarily model vehicle kinematics and traffic scenes, whereas indoor service robotics requires fine-grained object-centric reasoning (e.g., appliances, containers, manipulable items), multi-room navigation, and long-horizon task execution. In indoor contexts, prior work has explored robot--IoT communication for sensor fusion, system integration, and security~\cite{farnham2021umbrella, tashtoush2021enhancing, krzykowska2021secure, aroulanandam2022sensor, romeo2020internet}. Yet these systems typically do not study 
(1) how heterogeneous IoT observations can be fused into a shared semantic memory for downstream planning, or
(2) how such infrastructure signals change multi-robot coordination behavior beyond basic R2R sharing.
\textit{IndoorR2X} addresses this gap by introducing configurable indoor ``X'' sources (e.g., CCTV-derived object/location priors and device status reports) and evaluating how they affect coordination efficiency and reliability.

\subsection{LLM-Driven Planning and Coordination for Multi-Robot Systems}
LLMs are increasingly used to translate natural-language goals into structured plans, allocate sub-tasks across robots, and mediate multi-agent communication~\cite{li2025large, lykov2023llm, mandi2024roco, zhang2025lamma}. Several recent indoor or manipulation-centric systems use LLMs as operating-system-like coordinators or planners, combining perception, memory, and tool execution~\cite{chen2025emos, liu2025coherent, song2025collabot, zhu2025dexter}, and others explicitly compile LLM outputs into classical representations such as PDDL or behavior trees~\cite{kannan2024smart, bai2024twostep}.

A recurring limitation is that most LLM-based multi-robot frameworks assume that the ``information bottleneck'' is primarily robot-to-robot communication: the planner is fed only robots’ onboard observations (sometimes with simplified global state in simulation), and belief updates largely come from physical exploration and dialogue~\cite{ahn2024vader, ling2025elhplan, nayak2024long}. In contrast, \textit{IndoorR2X} explicitly models an R2X information channel by fusing robot observations with ambient IoT sensing into a global semantic state maintained by a coordination hub, allowing the LLM planner to reason over shared, time-stamped, cross-source state.

\subsection{Benchmarks Under Partial Observability and Multi-Robot Exploration}
Partial observability is central to embodied decision making and is commonly formalized through POMDP-style formulations~\cite{kaelbling1998planning}. In multi-robot settings, limited fields of view and incomplete maps make coordination challenging and often lead to redundant exploration, motivating classical work on coordinated exploration and frontier-based search~\cite{burgard2005coordinated, yamauchi1998frontier}.
Many embodied AI benchmarks, however, either expose near-global simulator state (implicitly giving planners oracle access) or focus on single-agent instruction following, making coordination effects difficult to measure~\cite{kant2022housekeep}. More recent LLM-centric benchmarks and systems do study multi-agent collaboration, but typically vary coordination protocols or planning abstractions without systematically evaluating IoT sensing as an external information source~\cite{huang2025compositional, lim2025dynamic, liu2024heterogeneous}.

\textit{IndoorR2X} complements these efforts by (i) enforcing realistic partial observability (each robot only knows its current view and visited areas) and (ii) providing configurable ``X'' sensing layouts, enabling controlled studies of how IoT-derived global context reduces redundant exploration and replanning cost.

\section{IndoorR2X Benchmark Design}
\label{sec:benchmark}
To rigorously evaluate multi-robot coordination in realistic indoor settings, we designed a benchmark featuring challenging scenarios that highlight the benefits of the R2X paradigm, where IoT sensors augment the capabilities of a robot fleet operating under practical perception constraints.
Our benchmark builds on the AI2-THOR engine~\cite{ai2thor}, leveraging 10 artist-curated homes from ArchitecTHOR~\cite{procthor} and 75 modular apartments (multi-cabins) from RoboTHOR~\cite{deitke2020robothor}. Together, these 85 multi-room environments provide the scale required for complex household tasks involving joint navigation and manipulation (ArchitecTHOR) as well as navigation-only tasks (RoboTHOR). As such, they constitute a robust testbed for evaluating high-level coordination and cooperative behaviors.

A key differentiator of our benchmark compared to prior LLM-driven multi-robot frameworks~\cite{kannan2024smart, bai2024twostep, chen2025emos} is its explicit modeling of realistic sensor limitations. We depart from the common assumption of an omniscient simulation with oracle-level scene knowledge. Instead, each robot's environmental knowledge is strictly limited to its previously visited areas and current field of view. This constraint prevents any single agent from acquiring complete global knowledge on its own, which offers an ideal testbed for our R2X hypothesis: integrating information from static IoT sensors can substantially reduce the need for exhaustive exploration, enabling more efficient task planning and execution.

\begin{algorithm}[h]
\footnotesize
\DontPrintSemicolon
\caption{IndoorR2X Coordination Framework}
\label{alg:r2x_planning_final}
\KwIn{Task $\mathcal{T}$, Goal $\mathcal{G}$, Fleet $\mathcal{R}$, IoT $\mathcal{D}$}
$\mathcal{W} \gets \text{InitState}(\mathcal{R}, \mathcal{D}), \Pi \gets \emptyset, \texttt{replan} \gets \texttt{true}, \texttt{fails} \gets 0$\;

\While{$\neg \text{GoalSatisfied}(\mathcal{W}, \mathcal{G}) \land \texttt{fails} < \text{MAX\_FAILS}$}{
    \tcc{1. Online R2X Fusion (Eq.~\ref{eq:data_fusion})}
    $\mathcal{W} \gets \mathcal{F}(\mathcal{W}, \text{Observe}(\mathcal{R} \cup \mathcal{D}))$\;

    \tcc{2. LLM-driven Planning (Eq.~\ref{eq:planning})}
    \If{$\texttt{replan} \lor \text{NeedsReplan}(\Pi, \mathcal{W})$}{
        $\text{HaltAffectedRobots}(\Pi, \mathcal{W})$ 
        $\Pi \gets f_{\text{LLM}}(\mathcal{P}(\mathcal{T}, \mathcal{W}))$\;
        \lIf{$\neg \text{IsValid}(\Pi, \mathcal{R})$}{$\texttt{fails}\texttt{++}$, \textbf{continue}}
        $\texttt{replan} \gets \texttt{false}$\;
    }

    \tcc{3. Parallel Action Dispatch}
    $V_{rdy} \gets \{v \in \Pi \mid \text{status}(v) = \text{PENDING} \land \text{DepsDone}(v)\}$\;
    $R_{idle} \gets \{r \in \mathcal{R} \mid S_r.\sigma = \text{IDLE}\}$\;

    \For{$v \in V_{rdy}$}{
        \lIf{$R_{idle} = \emptyset$}{\textbf{break}} 
        $r^* \gets \text{MatchRobot}(v, R_{idle}, \mathcal{W})$\;
        \If{$r^* \neq \texttt{None}$}{
            $\text{Dispatch}(r^*, v)$\;
            $S_{r^*}.\sigma \gets \text{EXECUTING}, \text{status}(v) \gets \text{RUNNING}$\;
            $R_{idle} \gets R_{idle} \setminus \{r^*\}$\;
        }
    }

    \tcc{4. Asynchronous Event Monitoring}
    $Evt \gets \text{Wait}(\mathcal{R} \cup \mathcal{D}, \Delta t)$\;
    
    \uIf{$Evt = \text{Timeout}$}{
        $\texttt{replan} \gets \texttt{replan} \lor \text{DetectStall}(\Pi, \mathcal{R})$\;
    }\uElseIf{$Evt = \text{IoTUpdate}$}{
        \lIf{$\text{Relevant}(Evt, \Pi)$}{$\texttt{replan} \gets \texttt{true}$}
    }\ElseIf{$Evt = \text{ActionDone}$}{
        $v \gets Evt.\text{act}, S_{Evt.\text{src}}.\sigma \gets \text{IDLE}$\;
        \uIf{$Evt.\text{res} = \text{SUCCESS}$}{
            $\text{status}(v) \gets \text{DONE}, \mathcal{W} \gets \text{ApplyEffects}(\mathcal{W}, v)$\;
            $\texttt{fails} \gets 0$\;
        }\Else{
            $\text{status}(v) \gets \text{FAILED}$\;
            $\texttt{replan} \gets \texttt{true}, \texttt{fails}\texttt{++}$\;
        }
    }
}
\end{algorithm}

This is where the ``X'' in R2X becomes critical. Our benchmark integrates IoT devices that naturally and efficiently expand the global knowledge base available to the robots. Specifically, we simulate indoor CCTV systems by deploying static, third-party cameras throughout the environment. We randomly configure the layout such that approximately 50\% of the house area is covered by CCTV, leaving the remaining space for the robots’ autonomous exploration. The video feeds from these cameras are processed by a vision-language model (VLM) (e.g., Qwen-VL~\cite{Qwen-VL}) to produce time‑stamped, text‑based event logs. These logs are continuously streamed to a central coordination hub, where they are fused with the robots’ onboard observations to maintain a global, real‑time belief state. With this enriched global state, an LLM‑based online planner can dynamically synthesize executable, parallel task plans for the multi‑robot team, which are then executed by embodied agents within the virtual environment.

\section{The IndoorR2X Framework}
\label{sec:indoorr2x}

The \textit{IndoorR2X} framework enables autonomous, multi-agent coordination through a continuous, online cycle of perception, planning, and action. A centralized \emph{coordination hub} maintains a global semantic state and mediates execution via an LLM-based planner and a parallel action orchestrator.

\subsection{Global Semantic State and R2X Data Fusion}
\label{subsec:world_model}

At the core of our framework is the \textbf{coordination hub}, which maintains a global semantic state $\mathcal{W}_t$ at time $t$ as a tuple of entity sets:
\begin{equation}
\label{eq:world_model}
\mathcal{W}_t = (\mathbf{S}^R_t, \mathbf{S}^O_t, \mathbf{S}^A_t),
\end{equation}
where $\mathbf{S}^R_t$, $\mathbf{S}^O_t$, and $\mathbf{S}^A_t$ denote the sets of robot, object, and area (room) states, respectively.

For each robot $r_i$ in the fleet $\mathcal{R}=\{r_1,\dots,r_N\}$, its state $S_{r_i}\in \mathbf{S}^R_t$ is:
\begin{equation}
\label{eq:robot_state}
S_{r_i} = (id_i, \mathbf{p}_i, \theta_i, \sigma_i, \text{inv}_i, \text{skills}_i),
\end{equation}
where $id_i$ is a unique identifier; $\mathbf{p}_i\in \mathbb{R}^3$ is position; $\theta_i\in [0,360)$ is yaw; $\sigma_i\in\{\text{IDLE},\text{EXECUTING},\text{CANCELING},\dots\}$ is status; $\text{inv}_i$ is the payload identifier (or \texttt{None}); and $\text{skills}_i$ is the set of action capabilities.

For each discovered object $o_j$, its state $S_{o_j}\in \mathbf{S}^O_t$ is:
\begin{equation}
\label{eq:object_state}
S_{o_j}=(id_j,type_j,\mathbf{p}_j,rec_j,\boldsymbol{\pi}_j,\text{room}_j,\text{src}_j,\tau_j),
\end{equation}
where $id_j$ is a unique identifier; $type_j$ is the class (e.g., \texttt{Microwave}, \texttt{Apple}); $\mathbf{p}_j\in\mathbb{R}^3$ is the last known position; $rec_j$ is the identifier of the parent receptacle containing the object (or \texttt{None} if uncontained), which governs visibility and reachability constraints; and $\boldsymbol{\pi}_j\in\{0,1\}^D$ is a binary vector encoding $D$ dynamic properties. This topological addition ensures the planner understands that an object's spatial coordinates ($\mathbf{p}_j$) are inaccessible if its parent receptacle is closed.

Let $\mathcal{P}_{\text{props}}$ be the property set with $D=|\mathcal{P}_{\text{props}}|$:
\begin{equation}
\label{eq:props}
\mathcal{P}_{\text{props}} = \{\texttt{isOpen}, \texttt{isToggled}, \texttt{isBroken}, \cdots \}.
\end{equation}
Thus, $\pi_{j,\texttt{isOpen}}=1$ indicates object $o_j$ is open. Finally, $\text{room}_j$ denotes the containing room, $\text{src}_j$ is the source (Robot ID or IoT Device ID), and $\tau_j$ is the timestamp of the last update.

A key innovation of our R2X approach is fusing heterogeneous observations from mobile robots and stationary IoT devices. The world model evolves via a transition function $\mathcal{F}$ that processes update messages $u_t$ from all agents (robots $\mathcal{R}$ and IoT devices $\mathcal{D}=\{d_1,\dots,d_M\}$):
\begin{equation}
\label{eq:data_fusion}
\mathcal{W}_{t+1} = \mathcal{F}(\mathcal{W}_t, u_t), \quad u_t \in \{\text{observations from } \mathcal{R}\cup\mathcal{D}\}.
\end{equation}
Specifically, the transition function $\mathcal{F}$ updates the topological hierarchy of $\mathcal{W}_t$ during object manipulation. When a robot $r_i$ successfully executes a \texttt{Pickup} on object $o_j$, $\mathcal{F}$ updates the robot's payload ($S_{r_i}.\text{inv}_i \gets id_j$) and assigns the robot as the object's new parent container ($S_{o_j}.rec_j \gets id_i$). Conversely, a \texttt{Put} action targeting receptacle $o_k$ clears the robot's payload and updates the object's parent to the target ($S_{o_j}.rec_j \gets id_k$). This rigorous bookkeeping ensures that the LLM planner's subsequent queries reflect the true nested state of the environment.

\subsection{Online R2X Planning and Execution}
\label{subsec:planning_execution}

Given a high-level task $\mathcal{T}$ with goal condition $\mathcal{G}$, the system repeats a \emph{sense--plan--act} loop until $\mathcal{G}$ is satisfied or no feasible progress remains. The online planner queries an LLM to produce a multi-agent plan represented as a dependency graph (DAG)
\begin{equation}
\label{eq:planning}
\Pi = (\mathcal{V}, \mathcal{E}) \gets f_{\text{LLM}}(\mathcal{P}(\mathcal{T}, \mathcal{W}_t)),
\end{equation}
where $\mathcal{P}(\mathcal{T}, \mathcal{W}_t)$ serializes the task, current world state, robot capabilities, and an output schema for the planner,
$\mathcal{V}$ is a set of action steps (nodes) and $\mathcal{E}$ encodes dependencies (edges). Each action node $v\in\mathcal{V}$ specifies an action type and parameters, plus execution constraints:
\begin{equation}
\label{eq:action_node}
v = (a, \text{params}, \text{req\_skills}, r_{\text{pref}}),
\end{equation}
where $\text{req\_skills}$ denotes mandatory capabilities (e.g., manipulation vs.\ pure navigation), and $r_{\text{pref}}$ is an optional targeted robot. During the parallel dispatch phase, the orchestrator dynamically assigns nodes with satisfied dependencies to available, idle robots. This runtime matching relies on state-aware heuristics—evaluating spatial proximity, current inventory status, and required camera horizon ($\phi$) adjustments—to maximize parallel execution across the fleet. Finally, an asynchronous \textbf{execution monitor} handles failures and state changes during embodied multi-agent execution. It actively polls for physical event resolutions (e.g., \texttt{ActionDone}, simulator collisions) and dynamic \texttt{IoTUpdate} broadcasts. Upon detecting task failures, insurmountable simulator stalls, or relevant environmental shifts, the monitor halts only robots whose current actions depend on the failed or invalidated plan nodes, while allowing unaffected robots to continue executing independent branches of the DAG. In our implementation, a global halt is used only as a conservative fallback when the dependency graph cannot isolate the affected branch safely.

\begin{table*}[h!]
\footnotesize
\setlength{\tabcolsep}{1.8pt}
\centering
\caption{\textbf{Overall Performance Evaluation.} Results with GPT-4.1 as the LLM planner in a three-robot setting.}
\label{tab:ablation_comm}
\vspace{-0.5em}
\begin{tabular}{lcccc}
\toprule
\textbf{Configuration} & \textbf{Success Rate} $\uparrow$ &  \textbf{Avg. Action Steps/Scene} $\downarrow$  & \textbf{Avg. Path Length/Scene} $\downarrow$ & \textbf{Avg. LLM Tokens/Scene} $\downarrow$ \\
\midrule
 \multicolumn{5}{l}{
\textcolor[gray]{0.7}{Comparison to prior works (adapted to our setting in Sec.~\ref{sec:benchmark}, e.g., target positions are unknown before exploration.)}}\\
SMART-LLM \textcolor[gray]{0.7}{(Adapted)}~\cite{kannan2024smart} (IROS 2024) & 88\% & 124 & 119 m & \underline{43,397}  \\
EMOS \textcolor[gray]{0.7}{(Adapted)}~\cite{chen2025emos} (ICLR 2025)  & 88\% & 135 & 122 m& 51,394  \\
 \hdashline
 \multicolumn{5}{l}{
\textcolor[gray]{0.7}{Our method (w/ ablations), which compares three communication configurations.}}\\
IR (w/o X \& Inter-robot Comm.)& 66\% & 186  & 137 m &54,572\\
R2R (w/o X Comm.) & \textbf{92\%} & \underline{116} & \underline{99 m} & 47,875\\
\textbf{R2X} & \textbf{92\%} & \textbf{108}  & \textbf{88 m} & \textbf{42,438}\\
\bottomrule
\end{tabular}
\end{table*}

\vspace{-0.5em}

\begin{table*}[h!]
\setlength{\tabcolsep}{1.2pt}
\centering
\caption{\textbf{Performance comparison across LLM sizes.} Results use the R2X configuration with three robots.}
\label{tab:ablation_LLMs}
\vspace{-0.5em}
\begin{tabular}{lcccc}
\toprule
\textbf{Model} & \textbf{Success Rate} $\uparrow$ &  \textbf{Avg. Action Steps/Scene} $\downarrow$  & \textbf{Avg. Path Length/Scene} $\downarrow$ & \textbf{Avg. LLM Tokens/Scene} $\downarrow$ \\
\midrule
Llama-3.1-8b-instruct (8 billion parameters) & 6\% &  47   & 55 m & 50,512\\
Gemma-3-27b-it (27 billion parameters) & 64\% &  113  &  94 m& 26,753\\
GPT-4.1 & 92\% & 108  & 88 m & 42,438\\
\bottomrule
\end{tabular}
\end{table*}

\subsection{Action Execution}
\label{subsec:low_level}

Abstract plan steps are realized by low-level executors that embed procedural knowledge to handle common failure modes. For example, executing \texttt{slice\_object} is a state-aware subroutine that checks inventory constraints, drops objects if needed, navigates to an appropriate workspace, and performs slicing with retries as applicable. All simulator or hardware interactions are routed through a fault-tolerant interface that sandboxes calls to prevent single-component failures from crashing the entire system.

\section{EXPERIMENTS AND RESULTS}
\label{sec:experiments}

We evaluate \textit{IndoorR2X} through a series of controlled ablation studies to isolate the impact of (i) information sharing mechanisms, (ii) LLM planner capability, and (iii) the reliability of the R2X channel. Unless otherwise stated, all experiments involve a team of three robots performing tasks from the suite described in Sec.~\ref{sec:benchmark}, utilizing the framework detailed in Sec.~\ref{sec:indoorr2x}.

\subsection{Experimental Setup}
We evaluate \textit{IndoorR2X} on the 85 virtual scenes and task suites detailed in Sec.~\ref{sec:benchmark}. To isolate the effects of information sharing, we compare three protocols: \textbf{IR (Isolated)}, where robots rely solely on local perception; \textbf{R2R}, where robots share a merged map; and \textbf{R2X (Ours)}, which augments R2R with real-time IoT updates (e.g., CCTV priors). To assess the system's sensitivity to reasoning capabilities, we evaluate three different LLMs as the central planner, keeping the perception and execution modules fixed. To test the robustness of the R2X integration, we systematically introduce artificial constraints during our ablations. These include varying the IoT communication latency ($t_{\text{delay}}$), scaling the robot team size ($N=2$ to $6$), and injecting perception failures (omission and corruption) into the infrastructure data stream.

\subsection{Evaluation Metrics}
We report \textbf{Success Rate (SR)} (percentage of fully completed episodes) and three efficiency metrics where lower values indicate better performance: \textbf{Avg. Action Steps} (cumulative low-level navigation and manipulation actions), \textbf{Avg. Path Length} (total meters traveled by the fleet), and \textbf{Avg. LLM Tokens} (proxy for planning cost).

\begin{figure*}[t]
  \centering
  \begin{minipage}[t]{0.55\linewidth}
    \centering
    \includegraphics[width=\linewidth]{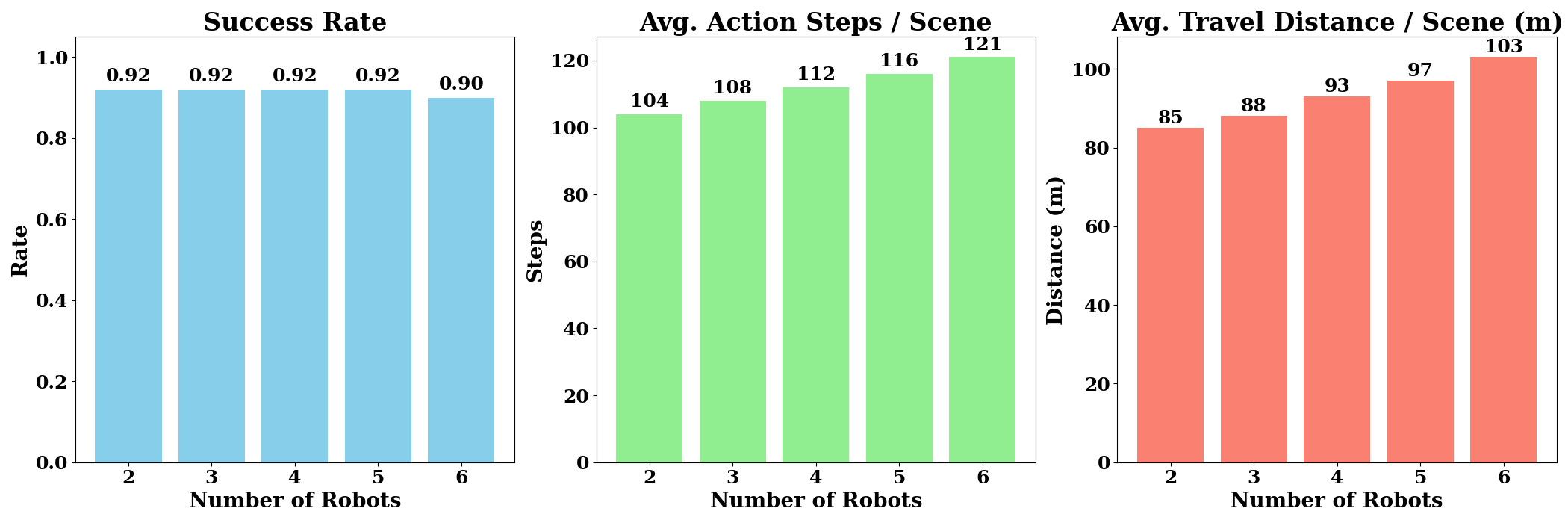}
    \vspace{-0.7em}
    \caption{\textbf{Scalability analysis.} Success rate (left) and efficiency metrics (center/right) as a function of team size ($N=2$ to $6$). While success remains stable up to $N=5$, the coordination overhead (total distance traveled) increases with fleet size.}
    \label{fig:robot_num}
  \end{minipage}\hfill
  \begin{minipage}[t]{0.43\linewidth}
    \centering
    \includegraphics[width=\linewidth]{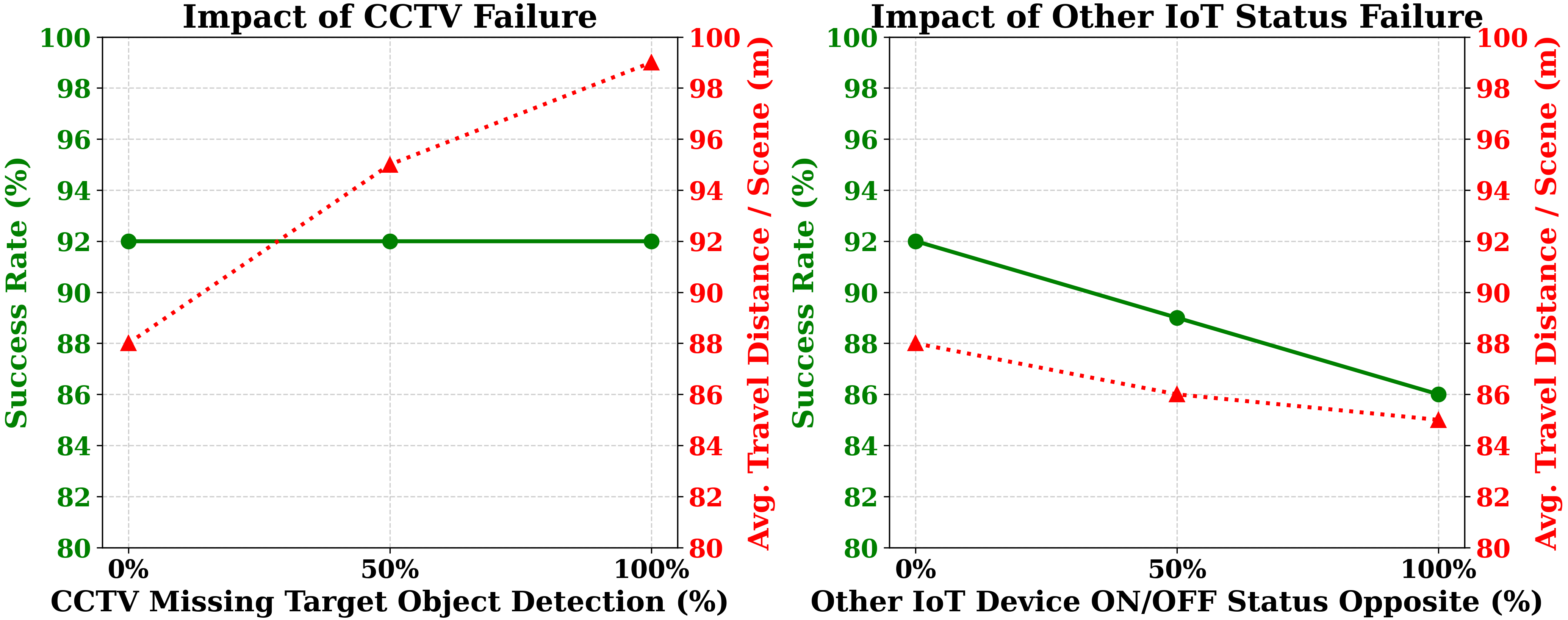}
    \vspace{-0.7em}
    \caption{\textbf{Robustness to ``X'' failures.} The system is resilient to missing detections (left), maintaining a constant success rate at the cost of increased travel. However, incorrect semantic status reports (right) significantly impact success, as false positives can lead to unrecoverable planning errors.}
    \label{fig:failure_analysis}
  \end{minipage}
\end{figure*}

\begin{table*}[th!]
\vspace{-1.5em}
\centering
\caption{\textbf{Effect of IoT latency on R2X performance.} Using GPT-4.1 as the LLM planner with R2X configuration, with three robots. }
\label{tab:iot_latency}
\vspace{-0.5em}
\setlength{\tabcolsep}{6pt}
\begin{tabular}{lcccc}
\toprule
\textbf{IoT Latency} & \textbf{Success Rate} $\uparrow$ &  \textbf{Avg. Action Steps/Scene} $\downarrow$  & \textbf{Avg. Path Length/Scene (m)} $\downarrow$ & \textbf{Avg. LLM Tokens/Scene} $\downarrow$ \\
\midrule
$t_{\text{delay}}$ = 0 (no delay)    & \textbf{92\%} & \textbf{108}  & \textbf{88 m} & \textbf{42,438}\\
$t_{\text{delay}}$ = 5               & \underline{90\%} & \underline{111} & \underline{95 m} & \underline{48,583} \\
$t_{\text{delay}}$ = 10              & 87\% & 116 & 101 m & 49,374\\      
\bottomrule
\end{tabular}
\end{table*}

\subsection{Impact of Communication Configuration}
Table~\ref{tab:ablation_comm} compares our approach with adapted state-of-the-art baselines and highlights the role of information sharing under partial observability. 
We adapt SMART-LLM and EMOS to our partial-observability setting by removing oracle access to object locations and device states before exploration. All methods use the same robot embodiment, action API, task suite, success criteria, and low-level executor. The adapted baselines receive only robot-side observations and inter-robot communication, without external IoT observations. Their planning prompts follow the original methods as closely as possible, with only action names and simulator interfaces mapped to our environment.
Compared to prior works that rely solely on R2R configurations (SMART-LLM and EMOS), our proposed R2R and R2X configurations improve overall task success by 4\%. Notably, our full R2X method is significantly more efficient than both baselines, reducing average path length by over 26\% compared to SMART-LLM and yielding the most token-efficient performance among all evaluated methods.

Within our ablations, the independent \textbf{IR} baseline struggles with a low success rate due to redundant exploration and uncoordinated actions. Enabling \textbf{R2R} communication boosts this success rate by nearly 40\% relative and reduces path length by $\sim$28\%. Incorporating IoT infrastructure data (\textbf{R2X}) maintains this high success rate while further improving execution efficiency: it reduces average action steps by $\sim$7\% and path length by an additional $\sim$11\% compared to R2R. This suggests that inter-robot sharing is important for task feasibility, while infrastructure sensing helps reduce unnecessary exploration. These spatial priors also reduce the planning burden on the LLM: R2X decreases average LLM token usage by over 11\% compared to R2R, which can translate to faster inference and lower operational cost.

\subsection{Impact of LLM Scale}
Table~\ref{tab:ablation_LLMs} analyzes the impact of the planner's model size. \textbf{GPT-4.1}~\cite{achiam2023gpt} achieves the highest reliability (92\% SR). \textbf{Gemma-3-27b}~\cite{team2025gemma} shows promise as a cost-effective alternative, achieving lower token footprint and respectable efficiency metrics, but suffering a drop in success rate (64\%). The smaller \textbf{Llama-3.1-8b}~\cite{dubey2024llama} fails in most episodes (6\% SR); note that its low action count is an artifact of early failure rather than efficiency. These results suggest a trade-off where smaller models may suffice for simpler sub-tasks, but capable frontier models are required for high-level coordination.

\subsection{Impact of IoT Latency}
We introduce artificial delays of $t_{\text{delay}}$ steps to the IoT stream before fusion (Table~\ref{tab:iot_latency}). Higher latency degrades performance: at $t_{\text{delay}}=10$, success rate drops to 87\%, and path length increases by $\sim$15\%. Stale IoT observations can cause outdated allocations, requiring replanning and extra travel. This suggests that the fusion module should be timestamp-aware, down-weighting or verifying old observations before using them as planning preconditions. Future systems can also reduce detail loss and computation with confidence-aware, event-triggered VLM updates, refreshing text logs only when motion or scene changes are detected while reusing cached semantic observations for unchanged regions.

\subsection{Scalability and Team Size}
We vary the robot team size from two to six agents (Fig.~\ref{fig:robot_num}). The success rate remains robust ($>$90\%) for teams of up to five robots, indicating effective conflict resolution. However, aggregate path length and action steps increase with team size, mainly due to spatial interference in narrow rooms, less stable task allocation among eligible robots, and larger joint-state/DAG prompts for the planner. This suggests that while IndoorR2X generalizes to teams larger than three robots, hierarchical or decentralized coordination will be important for larger fleets.

\subsection{Robustness to ``X'' Failures}
We evaluate system resilience against two types of IoT failures: \emph{omission} (missing detections) and \emph{corruption} (incorrect status). As shown in Fig.~\ref{fig:failure_analysis}, the system is highly robust to omission: even with 100\% of CCTV object detections missing, the success rate remains constant (92\%), though path length increases as robots are forced to actively explore.
In contrast, semantic corruption (e.g., reporting a device is ``OFF'' when it is ``ON'') is more detrimental, linearly reducing success rate. This asymmetry arises because missing information mainly delays planning, whereas false semantic information can make the planner skip necessary preconditions. Although our tests focus on omission and incorrect status reports, other VLM errors such as object misclassification, wrong-room localization, and missing receptacle relations may have different effects, motivating confidence-aware IoT fusion and robot-side verification.

\begin{figure*}[t]
  \centering
  \includegraphics[width=.99\linewidth]{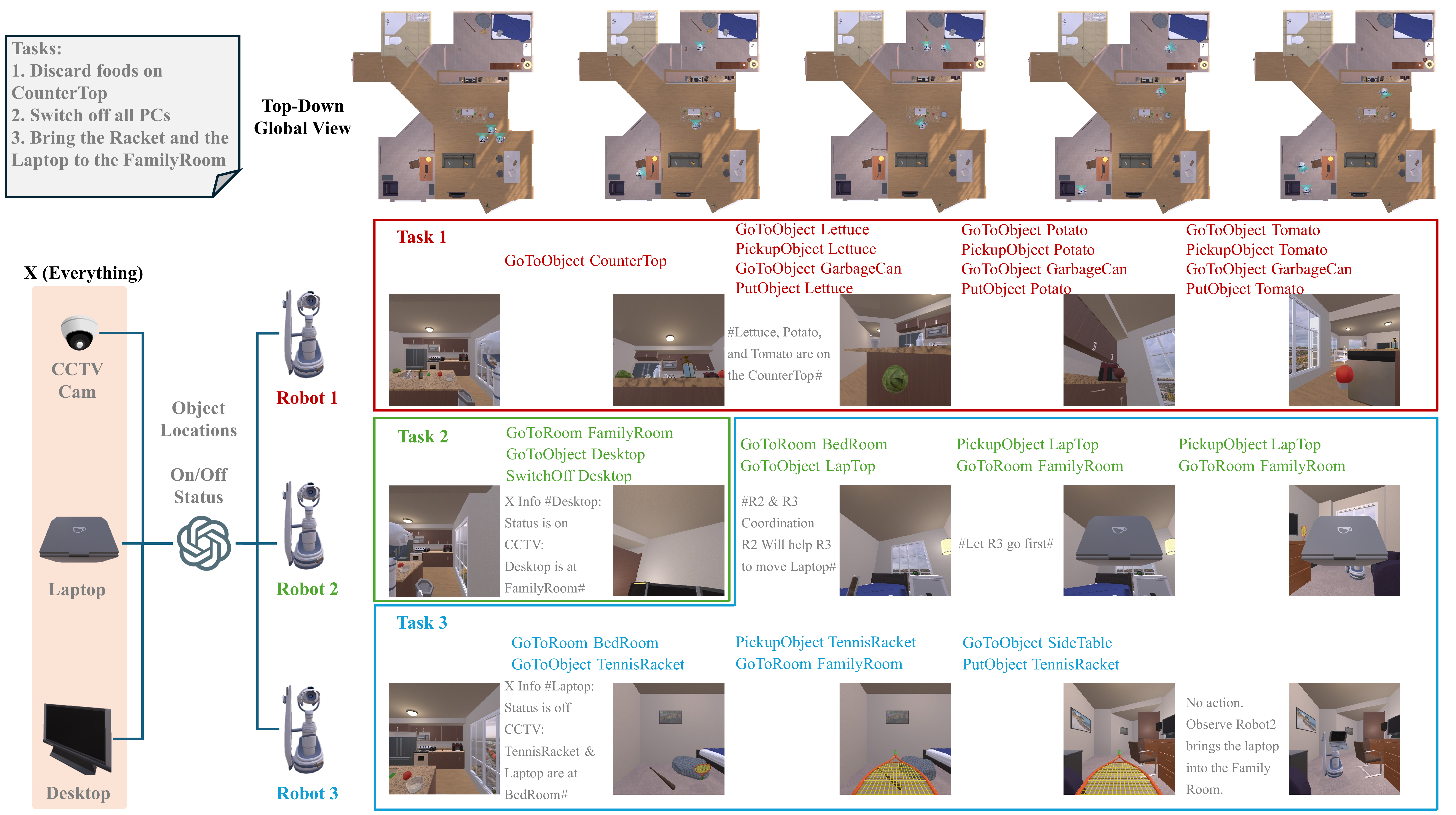}
  \vspace{-0.4em}
  \caption{\textbf{Qualitative demonstration of \textit{IndoorR2X} (simulation environment).} Three robots and IoT sensors coordinate to efficiently dispose of perishables, power down devices, and consolidate items in the family room.}
  \label{fig:qualitative_illustration}
\end{figure*}

\begin{figure*}[t]
  \centering
  \includegraphics[width=.97\linewidth]{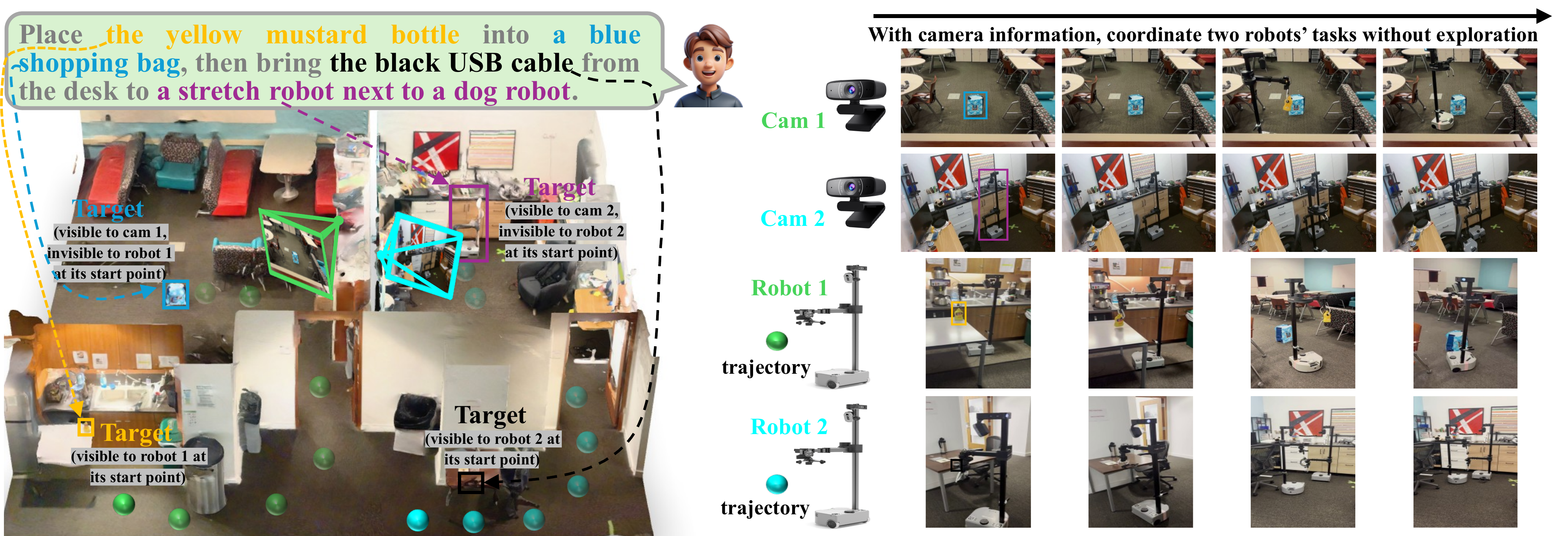}
  \vspace{-0.4em}
  \caption{\textbf{Illustration of our real-world experiment.} Two mobile Stretch robots jointly perform tasks in a three-room environment, utilizing two web cameras for out-of-sight visibility. A third Stretch robot stands stationary by a robot dog as a target.}
  \label{fig:real_test}
\end{figure*}

\subsection{Qualitative Analysis}
Fig.~\ref{fig:qualitative_illustration} visualizes a representative rollout involving three robots. The R2X integration allows the system to instantaneously populate the global semantic state with object locations (e.g., of the tennis racket and laptop) detected by CCTV.
Consequently, Robot 3 navigates directly to the tennis racket without a search phase, while Robot 2, upon finishing its task at the desktop, seamlessly transitions to assist Robot 3 with the laptop transport. The paths (shown in the top-down view) exhibit minimal overlap, demonstrating that the shared global context enables efficient spatial distribution of the fleet. This behavior qualitatively confirms our quantitative findings: R2X reduces search uncertainty, converting an exploration problem into a more efficient routing problem.

\subsection{Real-World Experiment} 
To validate our framework beyond simulation, we deploy \textit{IndoorR2X} in a physical environment. This real-world study utilizes Stretch robots~\cite{kemp2022design} and external web cameras, closely mirroring the sensing and embodiment configurations of our virtual trials. As shown in Fig.~\ref{fig:real_test}, the setup features a three-room environment. Two mobile Stretch robots are initialized in a room without line-of-sight to either of their targets: the blue shopping bag or the third, stationary Stretch robot positioned next to a dog robot. Traditional multi-robot coordination methods would require the mobile robots to exhaustively explore all rooms to locate these targets. By contrast, our approach broadens the global perception field using two web cameras. The previously unknown target locations are rapidly detected via Qwen-VL and localized within the shared environment representation. Consequently, \textit{IndoorR2X} enables the two mobile Stretch robots to largely bypass the exploration phase and navigate directly to the targets for manipulation and delivery. This reduces execution time and streamlines overall task completion.

\section{CONCLUSION}

We presented \textit{IndoorR2X}, a benchmark and framework for extending V2X-style infrastructure sensing to indoor multi-robot coordination. By fusing onboard robot perception with ambient IoT sensors such as CCTV cameras, \textit{IndoorR2X} constructs a shared global semantic state that helps mitigate partial observability. Our evaluations in simulation and initial physical deployments show that R2X integration reduces redundant exploration, path length, action steps, and LLM token cost while improving overall task efficiency. Our robustness analysis further shows that the system is resilient to missing sensor data but more sensitive to semantic corruption, highlighting the need for verification mechanisms. Future work should explore confidence-aware IoT fusion, event-triggered VLM updates, and hierarchical or decentralized coordination for larger robot fleets.

%%%%%%%%%%%%%%%%%%%%%%%%%%%%%%%%%%%%%%%%%%%%%%%%%%%%%%%%%%%%%%%%%%%%%%%%%%%%%%%%%

% \newpage
\vspace{-0.5em}
\bibliographystyle{IEEEtran}
\bibliography{IEEEabrv,bibtex}

@article{ai2thor,
  author={Eric Kolve and Roozbeh Mottaghi and Winson Han and
          Eli VanderBilt and Luca Weihs and Alvaro Herrasti and
          Daniel Gordon and Yuke Zhu and Abhinav Gupta and
          Ali Farhadi},
  title={{AI2-THOR: An Interactive 3D Environment for Visual AI}},
  journal={arXiv},
  year={2017}
}

@inproceedings{deitke2020robothor,
  title={Robothor: An open simulation-to-real embodied ai platform},
  author={Deitke, Matt and Han, Winson and Herrasti, Alvaro and Kembhavi, Aniruddha and Kolve, Eric and Mottaghi, Roozbeh and Salvador, Jordi and Schwenk, Dustin and VanderBilt, Eli and Wallingford, Matthew and others},
  booktitle={Proceedings of the IEEE/CVF conference on computer vision and pattern recognition},
  pages={3164--3174},
  year={2020}
}

@inproceedings{procthor,
  author={Matt Deitke and Eli VanderBilt and Alvaro Herrasti and
          Luca Weihs and Jordi Salvador and Kiana Ehsani and
          Winson Han and Eric Kolve and Ali Farhadi and
          Aniruddha Kembhavi and Roozbeh Mottaghi},
  title={{ProcTHOR: Large-Scale Embodied AI Using Procedural Generation}},
  booktitle={NeurIPS},
  year={2022},
  note={Outstanding Paper Award}
}

@INPROCEEDINGS{v2x_1,
  author={Yang, Yang and Yu, Haiyang and Fu, Xiang and Ren, Yilong and Zhao, Yanan and Shi, Yuqi},
  booktitle={2025 IEEE Intelligent Vehicles Symposium (IV)}, 
  title={A Dynamic Priority-Based Batch Verification Scheme for V2X Communication in Vehicular Networks*}, 
  year={2025},
  volume={},
  number={},
  pages={781-786},
  doi={10.1109/IV64158.2025.11097768}}

@article{yusuf2024vehicle,
  title={Vehicle-to-everything (V2X) in the autonomous vehicles domain--A technical review of communication, sensor, and AI technologies for road user safety},
  author={Yusuf, Syed Adnan and Khan, Arshad and Souissi, Riad},
  journal={Transportation Research Interdisciplinary Perspectives},
  volume={23},
  pages={100980},
  year={2024},
  publisher={Elsevier}
}

@article{li2022v2x,
  title={V2X-Sim: Multi-agent collaborative perception dataset and benchmark for autonomous driving},
  author={Li, Yiming and Ma, Dekun and An, Ziyan and Wang, Zixun and Zhong, Yiqi and Chen, Siheng and Feng, Chen},
  journal={IEEE Robotics and Automation Letters},
  volume={7},
  number={4},
  pages={10914--10921},
  year={2022},
  publisher={IEEE}
}

@inproceedings{xiang2024v2x,
  title={V2x-real: a largs-scale dataset for vehicle-to-everything cooperative perception},
  author={Xiang, Hao and Zheng, Zhaoliang and Xia, Xin and Xu, Runsheng and Gao, Letian and Zhou, Zewei and Han, Xu and Ji, Xinkai and Li, Mingxi and Meng, Zonglin and others},
  booktitle={European Conference on Computer Vision},
  pages={455--470},
  year={2024},
  organization={Springer}
}

@article{wang2023accidentgpt,
  title={Accidentgpt: Accident analysis and prevention from v2x environmental perception with multi-modal large model},
  author={Wang, Lening and Ren, Yilong and Jiang, Han and Cai, Pinlong and Fu, Daocheng and Wang, Tianqi and Cui, Zhiyong and Yu, Haiyang and Wang, Xuesong and Zhou, Hanchu and others},
  journal={arXiv preprint arXiv:2312.13156},
  year={2023}
}

@article{mahmud2025integrating,
  title={Integrating llms with its: Recent advances, potentials, challenges, and future directions},
  author={Mahmud, Doaa and Hajmohamed, Hadeel and Almentheri, Shamma and Alqaydi, Shamma and Aldhaheri, Lameya and Khalil, Ruhul Amin and Saeed, Nasir},
  journal={IEEE Transactions on Intelligent Transportation Systems},
  year={2025},
  publisher={IEEE}
}

@article{wu2025v2x,
  title={V2x-llm: Enhancing v2x integration and understanding in connected vehicle corridors},
  author={Wu, Keshu and Li, Pei and Zhou, Yang and Gan, Rui and You, Junwei and Cheng, Yang and Zhu, Jingwen and Parker, Steven T and Ran, Bin and Noyce, David A and others},
  journal={arXiv preprint arXiv:2503.02239},
  year={2025}
}

@inproceedings{kant2022housekeep,
  title={Housekeep: Tidying virtual households using commonsense reasoning},
  author={Kant, Yash and Ramachandran, Arun and Yenamandra, Sriram and Gilitschenski, Igor and Batra, Dhruv and Szot, Andrew and Agrawal, Harsh},
  booktitle={European Conference on Computer Vision},
  pages={355--373},
  year={2022},
  organization={Springer}
}

@article{liu2024heterogeneous,
  title={Heterogeneous embodied multi-agent collaboration},
  author={Liu, Xinzhu and Guo, Di and Zhang, Xinyu and Liu, Huaping},
  journal={IEEE Robotics and Automation Letters},
  volume={9},
  number={6},
  pages={5377--5384},
  year={2024},
  publisher={IEEE}
}

@article{li2025large,
  title={Large language models for multi-robot systems: A survey},
  author={Li, Peihan and An, Zijian and Abrar, Shams and Zhou, Lifeng},
  journal={arXiv preprint arXiv:2502.03814},
  year={2025}
}

@article{bai2024twostep,
  title={Twostep: Multi-agent task planning using classical planners and large language models},
  author={Bai, David and Singh, Ishika and Traum, David and Thomason, Jesse},
  journal={arXiv preprint arXiv:2403.17246},
  year={2024}
}

@inproceedings{kannan2024smart,
  title={Smart-llm: Smart multi-agent robot task planning using large language models},
  author={Kannan, Shyam Sundar and Venkatesh, Vishnunandan LN and Min, Byung-Cheol},
  booktitle={2024 IEEE/RSJ International Conference on Intelligent Robots and Systems (IROS)},
  pages={12140--12147},
  year={2024},
  organization={IEEE}
}

@inproceedings{
chen2025emos,
title={{EMOS}: Embodiment-aware Heterogeneous Multi-robot Operating System with {LLM} Agents},
author={Junting Chen and Checheng Yu and Xunzhe Zhou and Tianqi Xu and Yao Mu and Mengkang Hu and Wenqi Shao and Yikai Wang and Guohao Li and Lin Shao},
booktitle={The Thirteenth International Conference on Learning Representations},
year={2025},
url={https://openreview.net/forum?id=Ey8KcabBpB}
}

@article{ahn2024vader,
  title={Vader: Visual affordance detection and error recovery for multi robot human collaboration},
  author={Ahn, Michael and Arenas, Montserrat Gonzalez and Bennice, Matthew and Brown, Noah and Chan, Christine and David, Byron and Francis, Anthony and Gonzalez, Gavin and Hessmer, Rainer and Jackson, Tomas and others},
  journal={arXiv preprint arXiv:2405.16021},
  year={2024}
}

@article{ling2025elhplan,
  title={ELHPlan: Efficient Long-Horizon Task Planning for Multi-Agent Collaboration},
  author={Ling, Shaobin and Wang, Yun and Fan, Chenyou and Lam, Tin Lun and Hu, Junjie},
  journal={arXiv preprint arXiv:2509.24230},
  year={2025}
}

@article{lim2025dynamic,
  title={Dynamic Task Adaptation for Multi-Robot Manufacturing Systems with Large Language Models},
  author={Lim, Jonghan and Kovalenko, Ilya},
  journal={arXiv preprint arXiv:2505.22804},
  year={2025}
}

@article{song2025collabot,
  title={CollaBot: Vision-language guided simultaneous collaborative manipulation},
  author={Song, Kun and Ma, Shentao and Chen, Gaoming and Jin, Ninglong and Zhao, Guangbao and Ding, Mingyu and Xiong, Zhenhua and Pan, Jia},
  journal={arXiv preprint arXiv:2508.03526},
  year={2025}
}

@article{nayak2024long,
  title={Long-horizon planning for multi-agent robots in partially observable environments},
  author={Nayak, Sid and Morrison Orozco, Adelmo and Have, Marina and Zhang, Jackson and Thirumalai, Vittal and Chen, Darren and Kapoor, Aditya and Robinson, Eric and Gopalakrishnan, Karthik and Harrison, James and others},
  journal={Advances in Neural Information Processing Systems},
  volume={37},
  pages={67929--67967},
  year={2024}
}

@article{lykov2023llm,
  title={Llm-mars: Large language model for behavior tree generation and nlp-enhanced dialogue in multi-agent robot systems},
  author={Lykov, Artem and Dronova, Maria and Naglov, Nikolay and Litvinov, Mikhail and Satsevich, Sergei and Bazhenov, Artem and Berman, Vladimir and Shcherbak, Aleksei and Tsetserukou, Dzmitry},
  journal={arXiv preprint arXiv:2312.09348},
  year={2023}
}

@inproceedings{liu2025coherent,
  title={Coherent: Collaboration of heterogeneous multi-robot system with large language models},
  author={Liu, Kehui and Tang, Zixin and Wang, Dong and Wang, Zhigang and Li, Xuelong and Zhao, Bin},
  booktitle={2025 IEEE International Conference on Robotics and Automation (ICRA)},
  pages={10208--10214},
  year={2025},
  organization={IEEE}
}

@inproceedings{mandi2024roco,
  title={Roco: Dialectic multi-robot collaboration with large language models},
  author={Mandi, Zhao and Jain, Shreeya and Song, Shuran},
  booktitle={2024 IEEE International Conference on Robotics and Automation (ICRA)},
  pages={286--299},
  year={2024},
  organization={IEEE}
}

@article{huang2025compositional,
  title={Compositional Coordination for Multi-Robot Teams with Large Language Models},
  author={Huang, Zhehui and Shi, Guangyao and Wu, Yuwei and Kumar, Vijay and Sukhatme, Gaurav S},
  journal={arXiv preprint arXiv:2507.16068},
  year={2025}
}

@article{Qwen-VL,
  title={Qwen-VL: A Versatile Vision-Language Model for Understanding, Localization, Text Reading, and Beyond},
  author={Bai, Jinze and Bai, Shuai and Yang, Shusheng and Wang, Shijie and Tan, Sinan and Wang, Peng and Lin, Junyang and Zhou, Chang and Zhou, Jingren},
  journal={arXiv preprint arXiv:2308.12966},
  year={2023}
}

@article{sabri2018integrated,
  title={An integrated semantic framework for designing context-aware internet of robotic things systems},
  author={Sabri, Lyazid and Bouznad, Sofiane and Rama Fiorini, Sandro and Chibani, Abdelghani and Prestes, Edson and Amirat, Yacine},
  journal={Integrated Computer-Aided Engineering},
  volume={25},
  number={2},
  pages={137--156},
  year={2018},
  publisher={SAGE Publications Sage UK: London, England}
}

@article{wilson2019robot,
  title={Robot-enabled support of daily activities in smart home environments},
  author={Wilson, Garrett and Pereyda, Christopher and Raghunath, Nisha and De La Cruz, Gabriel and Goel, Shivam and Nesaei, Sepehr and Minor, Bryan and Schmitter-Edgecombe, Maureen and Taylor, Matthew E and Cook, Diane J},
  journal={Cognitive systems research},
  volume={54},
  pages={258--272},
  year={2019},
  publisher={Elsevier}
}

@inproceedings{farnham2021umbrella,
  title={Umbrella collaborative robotics testbed and iot platform},
  author={Farnham, Tim and Jones, Simon and Aijaz, Adnan and Jin, Yichao and Mavromatis, Ioannis and Raza, Usman and Portelli, Anthony and Stanoev, Aleksandar and Sooriyabandara, Mahesh},
  booktitle={2021 IEEE 18th Annual Consumer Communications \& Networking Conference (CCNC)},
  pages={1--7},
  year={2021},
  organization={IEEE}
}

@article{tashtoush2021enhancing,
  title={Enhancing robots navigation in internet of things indoor systems},
  author={Tashtoush, Yahya and Haj-Mahmoud, Israa and Darwish, Omar and Maabreh, Majdi and Alsinglawi, Belal and Elkhodr, Mahmoud and Alsaedi, Nasser},
  journal={Computers},
  volume={10},
  number={11},
  pages={153},
  year={2021},
  publisher={MDPI}
}

@article{krzykowska2021secure,
  title={Is Secure Communication in the R2I (Robot-to-Infrastructure) Model Possible? Identification of Threats},
  author={Krzykowska-Piotrowska, Karolina and Dudek, Ewa and Siergiejczyk, Miros{\l}aw and Rosi{\'n}ski, Adam and Wawrzy{\'n}ski, Wojciech},
  journal={Energies},
  volume={14},
  number={15},
  pages={4702},
  year={2021},
  publisher={MDPI}
}

@article{aroulanandam2022sensor,
  title={Sensor data fusion for optimal robotic navigation using regression based on an IOT system},
  author={Aroulanandam, Vijay Vasanth and Sherubha, P and Lalitha, K and Hymavathi, J and Thiagarajan, R and others},
  journal={Measurement: Sensors},
  volume={24},
  pages={100598},
  year={2022},
  publisher={Elsevier}
}

@article{romeo2020internet,
  title={Internet of robotic things in smart domains: Applications and challenges},
  author={Romeo, Laura and Petitti, Antonio and Marani, Roberto and Milella, Annalisa},
  journal={Sensors},
  volume={20},
  number={12},
  pages={3355},
  year={2020},
  publisher={MDPI}
}

@article{achiam2023gpt,
  title={Gpt-4 technical report},
  author={Achiam, Josh and Adler, Steven and Agarwal, Sandhini and Ahmad, Lama and Akkaya, Ilge and Aleman, Florencia Leoni and Almeida, Diogo and Altenschmidt, Janko and Altman, Sam and Anadkat, Shyamal and others},
  journal={arXiv preprint arXiv:2303.08774},
  year={2023}
}

@article{dubey2024llama,
  title={The llama 3 herd of models},
  author={Dubey, Abhimanyu and Jauhri, Abhinav and Pandey, Abhinav and Kadian, Abhishek and Al-Dahle, Ahmad and Letman, Aiesha and Mathur, Akhil and Schelten, Alan and Yang, Amy and Fan, Angela and others},
  journal={arXiv e-prints},
  pages={arXiv--2407},
  year={2024}
}

@article{team2025gemma,
  title={Gemma 3 technical report},
  author={Team, Gemma and Kamath, Aishwarya and Ferret, Johan and Pathak, Shreya and Vieillard, Nino and Merhej, Ramona and Perrin, Sarah and Matejovicova, Tatiana and Ram{\'e}, Alexandre and Rivi{\`e}re, Morgane and others},
  journal={arXiv preprint arXiv:2503.19786},
  year={2025}
}

@inproceedings{zhang2025lamma,
  title={Lamma-p: Generalizable multi-agent long-horizon task allocation and planning with lm-driven pddl planner},
  author={Zhang, Xiaopan and Qin, Hao and Wang, Fuquan and Dong, Yue and Li, Jiachen},
  booktitle={2025 IEEE International Conference on Robotics and Automation (ICRA)},
  pages={10221--10221},
  year={2025},
  organization={IEEE}
}

@article{gosrich2025online,
  title={Online multi-robot coordination and cooperation with task precedence relationships},
  author={Gosrich, Walker and Agarwal, Saurav and Garg, Kashish and Mayya, Siddharth and Malencia, Matthew and Yim, Mark and Kumar, Vijay},
  journal={IEEE Transactions on robotics},
  year={2025},
  publisher={IEEE}
}

@article{kaelbling1998planning,
  title={Planning and acting in partially observable stochastic domains},
  author={Kaelbling, Leslie Pack and Littman, Michael L and Cassandra, Anthony R},
  journal={Artificial Intelligence},
  volume={101},
  number={1--2},
  pages={99--134},
  year={1998},
  publisher={Elsevier}
}

@inproceedings{burgard2005coordinated,
  title={Coordinated multi-robot exploration},
  author={Burgard, Wolfram and Moors, Mark and Stachniss, Cyrill and Schneider, Frank E},
  booktitle={IEEE Transactions on Robotics},
  volume={21},
  number={3},
  pages={376--386},
  year={2005},
  publisher={IEEE}
}

@inproceedings{yamauchi1998frontier,
  title={Frontier-based exploration using multiple robots},
  author={Yamauchi, Brian},
  booktitle={Proceedings of the Second International Conference on Autonomous Agents},
  pages={47--53},
  year={1998}
}

@inproceedings{zhu2025dexter,
  title={DEXTER-LLM: Dynamic and Explainable Coordination of Multi-Robot Systems in Unknown Environments via Large Language Models},
  author={Zhu, Yuxiao and Chen, Junfeng and Zhang, Xintong and Guo, Meng and Li, Zhongkui},
  booktitle={2025 IEEE/RSJ International Conference on Intelligent Robots and Systems (IROS)},
  pages={10182--10189},
  year={2025},
  organization={IEEE}
}

@inproceedings{kemp2022design,
  title={The design of stretch: A compact, lightweight mobile manipulator for indoor human environments},
  author={Kemp, Charles C and Edsinger, Aaron and Clever, Henry M and Matulevich, Blaine},
  booktitle={2022 International Conference on Robotics and Automation (ICRA)},
  pages={3150--3157},
  year={2022},
  organization={IEEE}
}

\end{document}